\documentclass{article}
\usepackage{arxiv}
\usepackage{amsfonts,booktabs,multirow,lineno,hyperref,xcolor,amsmath}
\usepackage{blkarray, bigstrut,pifont}
\usepackage{booktabs}

\usepackage[utf8]{inputenc} 
\usepackage[T1]{fontenc}    
\usepackage{hyperref}       
\usepackage{url}            
\usepackage{booktabs}       
\usepackage{amsfonts}       
\usepackage{nicefrac}       
\usepackage{microtype}      
\usepackage{lipsum}
\usepackage{graphicx}
\graphicspath{ {./images/} }

\usepackage{amsmath,amssymb,bm}
\usepackage{adjustbox}
\usepackage{caption}
\usepackage{subfigure}
\usepackage{multirow}
\usepackage[english]{babel}
\usepackage{amsthm}

\title{One-step learning algorithm selection for classification via convolutional neural networks}

\author{
Sebasti\'an Maldonado\\
Department of Management Control and Information Systems \\
School of Economics and Business\\
University of Chile, Santiago, Chile\\
\texttt{sebastianm@fen.uchile.cl}
\And
Carla Vairetti \\
Universidad de los Andes \\
Chile \\
Facultad de Ingenier\'{i}a y Ciencias Aplicadas
\texttt{cvairetti@uandes.cl}
\And
Ignacio Figueroa \\
Universidad de los Andes \\
Chile \\
Facultad de Ingenier\'{i}a y Ciencias Aplicadas
}

\begin{document}
\maketitle

\begin{abstract}
As with any task, the process of building machine learning models can benefit from prior experience.
Meta-learning for classifier selection leverages knowledge about the characteristics of different datasets and/or the past performance of machine learning techniques to inform better decisions in the current modeling process.
Traditional meta-learning approaches first collect metadata that describe this prior experience and then use it as input for an algorithm selection model.
In this paper, however, a one-step scheme is proposed in which convolutional neural networks are trained directly on tabular datasets for binary classification.
The aim is to learn the underlying structure of the data without the need to explicitly identify meta-features.
Experiments with simulated datasets show that the proposed approach achieves near-perfect performance in identifying both linear and nonlinear patterns, outperforming the conventional two-step method based on meta-features.
The method is further applied to real-world datasets, providing recommendations on the most suitable classifiers based on the data's inherent structure
\newline
\footnote{NOTICE: This is a preprint of a published work. Changes resulting from the publishing process, such as editing, corrections, structural formatting, and other quality control mechanisms may not be reflected in this version of the document. Please cite this
work as follows: Maldonado, S., Vairetti, C., Figueroa, I. (2025) One-step learning algorithm selection for classification via convolutional neural networks. Information Sciences 721, 122610. DOI: https://doi.org/10.1016/j.ins.2025.122610.}. 
\end{abstract}

\section{Introduction}

Meta-learning is a broad field of machine learning that originated decades before the rise of deep learning. Some previous studies in this field have learned from task properties, constructing \emph{meta-features} that represent the datasets. The goal is then to transfer insights from the most similar tasks to a new task \cite{de2024meta,vettoruzzo2024advances}. The reasoning behind meta-learning has been extended to other relevant challenges associated with deep neural networks, such as transfer learning or multitask learning \cite{lughofer2022transfer,XIA2023119008}. However, these challenges are not related to this paper, which focuses on learning algorithm selection for classification \cite{AliS1,8951014}.

This paper proposes an approach for learning meta-features directly from tensor representations of data using convolutional neural networks (CNNs). Instead of using images as input, we use simulated tabular datasets with distinguishable linear and nonlinear patterns, framing it as a multiclass classification task. The main contributions are the following:

\begin{itemize}
    \item A novel meta-learning approach is proposed that uses deep learning to extract meta-features from simulated tabular data, avoiding manual feature engineering.
To the best of our knowledge, this is the first time such an approach has been proposed.
In previous algorithm selection for classification approaches, meta-features were designed manually via feature engineering, and then used as inputs for a learning machine \cite{AliS1,8951014}.
    \item A pre-trained model capable of recommending suitable classifiers for tabular datasets without retraining or fine-tuning is introduced.
Although deep learning has succeeded in tasks such as computer vision and text analytics, there are many applications where classification is usually performed on tabular datasets.
Medicine and business analytics are examples of research fields that rely on traditional data to make predictions and decisions via machine learning (see, e.g., \cite{deeva2025select,do2022early}).
    \item The proposed method significantly reduces the time required for model comparison, making it practical for real-world applications in fields like medicine and business analytics.
Suggesting suitable classifiers can alleviate the need for extensive model comparisons, which can be very time-consuming \cite{AliS1}.
    \item This work opens a new research direction in algorithm selection for classification using representation learning, contributing to the broader field of meta-learning.
\end{itemize}

This study seeks to answer the following questions: (1) Can linear and nonlinear data structures be learned from raw tabular datasets using deep learning?
and (2) If so, can this knowledge be leveraged to make suggestions on the most suitable machine learning method or methods for a specific task?
The study is consistent with the reasoning behind the ``No Free Lunch'' (NFL) theorem in the sense that it may not be effective to gain knowledge from tasks to apply it to a completely unrelated dataset.
However, existing meta-learning studies have shown the advantages of learning from prior experience \cite{hospedales2021meta}.
This paper is organized as follows.
Section \ref{sec:prior} provides an overview of relevant meta-learning studies.
The proposed meta-learning framework is presented in Section \ref{sec:Prop}.
Section \ref{sec:Exp} discusses the results obtained using simulated and benchmark datasets.
Finally, Section \ref{sec:Conc} summarizes the primary conclusions and presents possible directions for future developments.

\section{Prior Work on Meta-learning for Classifier Selection}\label{sec:prior}

Meta-learning is an umbrella term that encompasses several approaches that gain experience across tasks.
The goal of meta-learning is to avoid time-consuming ``trial and error'' processes and make better choices when estimating machine learning models.
Some examples of these approaches include the following (see \cite{hospedales2021meta} for a comprehensive literature review):

\begin{itemize}
\item \emph{Learning from previous model estimations and evaluations:} Given a set of possible configurations $\Theta$, the goal of this approach is to train a meta-learner based on previous model evaluations to make recommendations on a new task.
In a hyperparameter search task, for example, the accuracy of a classifier can be estimated on different datasets (or variants of the same datasets using bootstrapping or cross-validation), and a meta-learner can be used to suggest an optimal hyperparameter configuration (or a set of candidate configurations, called a \emph{portfolio} \cite{hospedales2021meta}).
Note that a configuration can be a set of hyperparameters, network architectures, or pipeline components \cite{hospedales2021meta}.
\item \emph{Transfer learning:} The reasoning behind this approach is to consider models that are trained with data from one or more sources as a starting point for developing a new model on a similar task.
Although this idea has been applied to traditional machine learning methods, this approach has been particularly successful in deep learning (DL) \cite{hospedales2021meta}.
In tasks such as object recognition and natural language processing (NLP), DL requires a large amount of data to achieve superior performance compared to other methods.
A \emph{pretrained model} can be constructed with publicly available data (e.g., Wikipedia or books in the case of NLP tasks \cite{GONZALEZ202158} or ImageNet in the case of visual object recognition \cite{MORID2021104115}), which is then adapted to the new task via \emph{fine-tuning}.
\item \emph{Learning from meta-features:} The goal of this approach is to make better decisions during machine learning by learning from meta-features or properties that describe the datasets.
These characterizations can be used for hyperparameter selection \cite{MU2022344} or classifier selection \cite{AliS1,8951014}, among other tasks.
A task similarity metric can be defined to transfer knowledge from one task to a new similar task, or a meta-learner can be constructed \cite{JMLR:v21:19-348}.
For example, in \cite{AliS1}, a decision tree learner is trained on several meta-features that were constructed on more than 100 datasets from the UCI Repository \cite{blake2015uci} and other sources.
A multiclass classification problem with five classes related to five different classification algorithms is defined.
Alternatively, information on dataset characteristics can be useful to infer the performance of feature selection methods \cite{ORESKI2017109}.
\end{itemize}

Algorithm selection in machine learning can be seen as a targeted subtask of AutoML.
AutoML refers to automated systems that aim to optimize the entire machine learning pipeline, including algorithm selection, hyperparameter tuning, feature engineering, and model evaluation.
The goal is to provide an end-to-end solution that minimizes the need for expert intervention, often treating the problem as a black box \cite{SALEHIN202452}.
In contrast, the present work focuses specifically on the algorithm selection problem.
Instead of searching blindly, the aim is to learn patterns about which ML algorithms work best for which kinds of datasets, using meta-learning.
It avoids the costly search process of full AutoML by leveraging prior knowledge.
The method proposed in this study is related to the latter approach in the sense that learning is performed from characterizations of other tasks to make recommendations for suitable classifiers.
However, instead of using a two-step approach by first constructing meta-features and then estimating a meta-learner, the proposed meta-learner is fed with tabular datasets directly.
Different purposes can be distinguished in the creation of meta-features in the sense that specific measures aim to identify specific patterns, such as feature interdependence, class overlap, or task similarity:

\begin{itemize}
\item \emph{Feature normality and dispersion:} Statistical measures used to describe a distribution can be considered to assess normality in the covariates.
Some common examples are skewness $\frac{{E(X - \mu)}^3}{\sigma^3}$ and kurtosis ($\frac{{E(X - \mu)}^4}{\sigma^4}$), with $\mu$ and $\sigma$ being the mean and standard deviation of variable $X$, respectively \cite{vijithananda2022feature}.
Other measures that can be related to feature normality are the interquartile range (IQR, the difference between quartile 3 and quartile 1) and the value of the 90th quantile \cite{AliS1}.
Other statistics that can assess tendency and dispersion are the arithmetic mean, the geometric mean, the harmonic mean, the trimmed mean, the standard deviation, and the mean absolute deviation (MAD, $\frac{\sum\left|x_i - \mu \right|}{N}$, with $i=1,...,N$ representing the samples) \cite{AliS1}.
Finally, the Index of Dispersion (ID) indicates whether the data points are scattered or clustered \cite{AliS1}.
\item \emph{Complexity:} Simple measures can be an indicator of the complexity of the learning task, such as the size of the dataset (number of rows and columns) and the number of classes.
These measures can be related to the expected training times \cite{michie1994machine}.
Alternatively, the percentage of missing values and outliers can be related to the complexity of the preprocessing step \cite{AliS1,rousseeuw2011robust}.
\item \emph{Feature interdependence:} The level of redundancy in a dataset can be assessed using the Pearson correlation ($\rho =\frac{\sigma_{XY}}{\sigma_X\sigma_Y}$) or by analyzing the eigenvalues that result from applying the principal component analysis (PCA) method for feature extraction \cite{michie1994machine}.
Some PCA-based meta-features are canonical correlations (square root of the eigenvalues), the first and last PC \cite{AliS1}, and the skewness and kurtosis of the first PC \cite{feurer2014using}.
\item \emph{Class overlap and feature relevance:} The meta-learning literature has reported different metrics to evaluate overall feature relevancy, such as the center of gravity (Euclidean norm between minority and majority classes) \cite{AliS1}, the entropy of classes ($H(C) = -\sum_{c}{\pi_c\log{\pi_c}}$), the mean mutual information (MMI), the equivalent number of variables (ENV, ratio between the class entropy and the MMI), and the noise-signal ratio ($NSR = \frac{\bar{H}(X) - \bar{M}(C,X)}{\bar{M}(C,X)}$) \cite{AliS1}.
A large value for the latter metric suggests that the dataset contains several irrelevant/noisy variables, and therefore, its dimensionality can be reduced without affecting the classification accuracy \cite{AliS1}.
\item \emph{Landmarks:} This strategy computes measures designed to assess task similarity, using the learners themselves to describe the dataset.
The idea is to compare differences in terms of performance between configurations and/or datasets using simple classifiers, such as 1-NN or naive Bayes \cite{gabbay2021isolation}.
\end{itemize}

It is important to note that, although meta-learning has usually been applied for algorithm recommendation in classification \cite{Wang2014,ZHU2018171}, other machine learning tasks in which meta-learning has shown good results include regression \cite{Lorena2018}, time series analysis \cite{LEMKE20102006}, data stream mining \cite{ROSSI201452}, and clustering \cite{gabbay2021isolation}.
The proposed meta-learning approach is discussed in the following section.
\section{Proposed Classifier Selection Framework via CNNs}\label{sec:Prop}

This paper proposes a novel meta-learning framework in which tabular datasets are introduced into a learning model directly, which differs from the traditional two-step approach that involves the construction of meta-features.
In this sense, each tabular dataset is treated as an ``image'', using CNNs for model training.
The reasoning behind this approach is to avoid the loss of information that occurs when meta-features are constructed and subsequently used as inputs for a traditional classifier, such as decision trees \cite{AliS1}.
The framework consists of four steps:
\begin{enumerate}
\item \textbf{Simulation of patterns:} Different training patterns are defined that can be related to the performance of learning models.
These patterns can be simulated under different noise and class overlap conditions, and then used to feed a CNN in a supervised manner.
For example, a logistic regression or a linear support vector machine (SVM) can be recommended when dealing with tabular datasets with a linear pattern.
The most suitable classifier is a hyperplane, while kernel-based SVM, NNs, or random forests can be recommended when datasets at hand follow a nonlinear pattern.
For a setting in which all tabular datasets are of similar size, each input consists of tuples in the form $\{({\bf x}_1,y_1),\ldots,({\bf x}_N,y_N)\}$, where ${\textbf x}_i \in \mathbb{R}^{P}$ and $y_i \in \{-1,1\}$ for $i=1,\ldots,N$.
In other words, the focus is limited to tabular datasets for binary classification tasks, but the approach can be extended easily to multiclass classification.
For each simulated pattern $c=1,\ldots,C$, $M$ different simulated datasets are considered.
Therefore, the dimensions of the input tensor are $N \times (P+1) \times (C \cdot M)$.
This setting with homogeneous datasets is referred to as CNN1.
Alternatively, a setting with datasets that are heterogeneous in the number of samples and variables can be simulated.
This setting is more compatible with meta-learning because the goal is to transfer the knowledge gained from simulated data to real-world datasets, which are heterogeneous in size.
Furthermore, such a setting allows the introduction of noise into the patterns in the form of irrelevant variables and/or samples that do not follow the simulated pattern, thus mitigating the risk of overfitting.
Similar to CNN1, $M$ different simulated datasets for each of the $C$ patterns are considered.
For each simulated dataset $mc=1,\ldots,MC$, $N_{mc}$ and $P_{mc}$ represent the sample size and number of covariates, respectively.
This setting with heterogeneous datasets is referred to as CNN2.
The first step is arguably the most challenging because only a comprehensive set of patterns under different conditions of noise would lead to an adequate application on real-world datasets.
The success of the approach strongly relies on this step.
As a first attempt, five different patterns that are well known in the machine-learning literature, such as the two moons or XOR patterns \cite{pmlr-v119-yamada20a}, are discussed.
It is believed that this paper opens an exciting new line of research in which different sets of simulated patterns can be designed.
\item \textbf{Dealing with inputs of different sizes:} Because the proposed goal is to make recommendations for real-world tabular datasets, it is expected that datasets will all be of different sizes.
Therefore, the training and test samples must be reshaped to a homogeneous size.
This is only required for the CNN2 setting.
Simple approaches are proposed for dealing with images of different sizes with CNNs, such as padding and principal component analysis (PCA).
Padding is used to resize tabular datasets that are smaller than a target size by extending the area over which a CNN processes.
This process is performed by introducing additional rows or columns of zeros \cite{Aggarwal18}.
PCA can, in contrast, reduce tabular datasets to a target size by finding (orthogonal) linear combinations of the original variables in such a way that the variance is maximized.
Thus, the datasets can be shrunk while keeping their inherent structure \cite{LIU2022109715}.
Under the CNN2 setting, there are heterogeneous datasets $\{\textbf{X}_{mc},\textbf{y}_{mc}\}_{mc=1}^{MC}$, with $\textbf{X}$ and $\textbf{y}$ representing the covariate matrix and target vector.
In order to define a common dimensionality $P$, padding or PCA can be used.
In the case of padding, the largest $P{mc}$ for all the simulated datasets ($P=\max_{mc}(P_{mc})$) can be selected and the missing dimensions filled with zeros for all the remaining datasets.
In the case of PCA, a target dimensionality $P$ can be defined, and $P$ components are selected for each dataset.
To define a common sample size $N$, padding or random undersampling can be used.
In the case of padding, the largest $N_{mc}$ for all the simulated datasets ($N=\max_{mc}(N_{mc})$) can be selected and the missing samples filled with zeros for all the remaining datasets.
In the case of random undersampling, the minimum sample size can be defined ($N=\min_{mc}(N_{mc})$), discarding observations randomly in all the datasets until this number is reached.
These two strategies can also be combined. After these preprocessing strategies, the input tensor for the next step has dimensions $N \times (P+1) \times (C \cdot M)$.
\item \textbf{CNN training:} Often used for image-related tasks such as pattern recognition, segmentation, or object detection, CNNs are among the most popular deep learning variants \cite{xing2017deep}.
CNNs are employed in this framework because they naturally handle tensor data and eliminate the need for separate feature extraction, which contributes to their superior performance in image recognition \cite{karatzoglou2020applying, zhao2019object}.
Note that $\{\textbf{X}_{mc},\textbf{y}_{mc}\}_{mc=1}^{MC}$ represent the inputs for the CNN model and are treated as images.
The labels for these ``images'' are a vector of size $MC$ indicating each of the $C$ different patterns simulated in step 1. This vector is referred to as $\textbf{z}$.
It is important to make this distinction because the vectors $\textbf{y}_{mc}$, which are the labels of the tabular datasets, can be confused with the labels of the tabular datasets.
The classifier trained with simulated data is referred to as $\mathcal{C}$.
\item \textbf{Application to real-world datasets:} The final step consists of applying the CNN model $\mathcal{C}$ to real-world datasets to identify one of the simulated patterns that were considered during training.
The pattern found can be linked to one or more suitable classifiers for this pattern.
It is important to assess the confidence of the classifier in its choice;
therefore, the probabilistic output of the network must be analyzed using a softmax function.
For example, a learning task with five simulated patterns is considered in our experimental design.
If the largest predicted probability for a given real-world dataset is 0.3, then the model is undecided, and no recommendation can be made because this probability is too close to a random choice (0.2 for a 5-class task).
In contrast, the model is certain when the probability of a given class is near 1, leading to a trustworthy recommendation.
Note that step 2 is required to preprocess the real-world datasets.
This preprocessing should be consistent with the one performed for the training datasets;
i.e., the real-world datasets should also have input sizes $N$ and $P$.
Assuming that there are $T$ real-world datasets, each (heterogeneous) input is referred to as $\{\textbf{X}^r_{t},\textbf{y}^r_{t}\}_{t=1}^{T}$.
After preprocessing, the dimensionality of the tensor for the application of the model is $N \times (P+1) \times T$.
Notice that assigning patterns $\textbf{z}^t$ to the real-world datasets is not the ultimate goal of the proposed framework;
the goal is to recommend one or more classifiers. To this end, the performance of various classifiers for tabular datasets with the simulated data is evaluated and the best ones are identified.
Once a pattern is assigned to a real-world dataset, the classifier that performs best for this pattern is recommended.
For example, if a random forest performs best for pattern $C=2$ and the classifier $\mathcal{C}$ assigns this pattern to the real-world dataset $t=1$ ($z_1^t=2$), then the random forest is recommended as the binary classifier for this particular task.
In contrast to the first stage of the framework, which follows a standard supervised scheme, a tailored strategy needs to be defined to evaluate the success of the transfer learning approach.
For this stage, the aim is to assign labels to real-world tabular datasets that do not necessarily follow a clear linear/nonlinear pattern.
The assumption is that these labels can be linked to classifiers that can perform well for these patterns.
\end{enumerate}

\section{Experimental Results}\label{sec:Exp}

To validate the proposed meta-learning framework, CNN models were first trained with simulated data exhibiting different linear and nonlinear patterns and their performances were compared with the traditional approach based on 
meta-features proposed in \cite{AliS1}.
The results of these experiments are shown in Section \ref{sec:expSIM}.
The application of CNNs with real-world tabular datasets is reported in Section \ref{sec:expBench}.
Finally, Section \ref{sec:expDisc} provides a discussion that summarizes the main experimental findings.
\subsection{Construction of the Classifier Selection Model}\label{sec:expSIM}

A total of 50,000 tabular datasets were simulated with five different patterns ($C=5$, $M=10,000$ datasets each).
Five two-dimensional patterns that act as classes for the meta-learner were defined: a linear pattern (C1), an XOR pattern (C2), a two-moons pattern (C3), a ``sandwich'' pattern (C4), and a quadratic pattern (C5).
The following two-dimensional patterns were defined:

\begin{itemize}
\item \textbf{Class 1 - linear pattern:} Two Gaussian functions are combined, one for each class, varying the mean and standard deviation of each Gaussian to create different overlap and noise conditions.
\item \textbf{Class 2 - XOR pattern:} Four Gaussian functions are combined, two for each class, varying the mean and standard deviation of each Gaussian to create different overlap and noise conditions.
The four Gaussians shape the well-known XOR pattern related to the exclusive disjunction logical operation.
This nonlinear pattern is also known as ``checkerboard''.
\item \textbf{Class 3 - ``Two moons'' pattern:} This well-known synthetic dataset constructs a swirl pattern shaped like two moons.
The conditions are varied in terms of noise and overlap.
\item \textbf{Class 4 - ``Sandwich'' pattern:} Samples are simulated using one Gaussian function interspersed with two other Gaussians related to a second class.
This results in a nonlinear pattern in the form of a ``sandwich''.
The mean and standard deviation of each Gaussian distribution are varied to create different overlap and noise conditions.
\item \textbf{Class 5 - Quadratic pattern:} Samples are simulated based on two quadratic functions, one for each class, with a marginal overlap.
The conditions are varied in terms of noise and overlap.
\end{itemize}

\begin{figure}[ht!]
\begin{center}
\subfigure[]{%
 \label{Fig1:first}
 \includegraphics[width=0.5\textwidth]{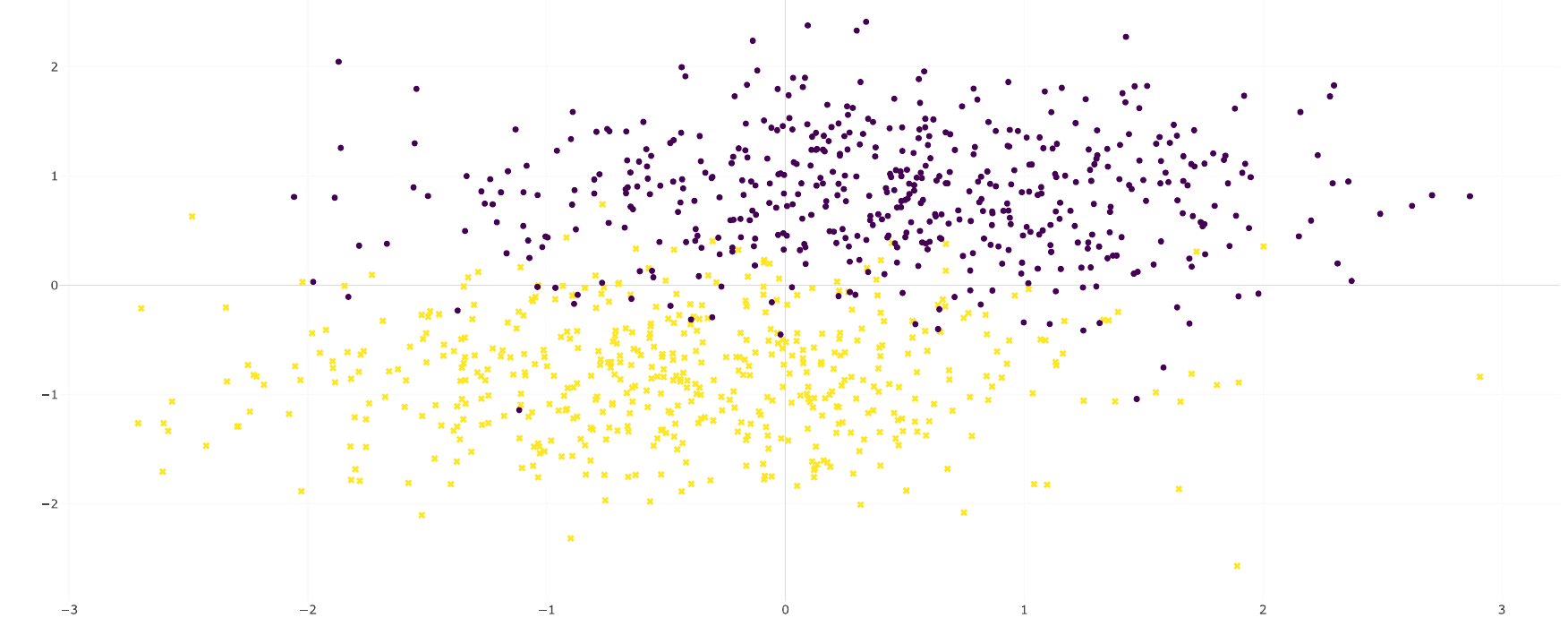}
 }%
 \subfigure[]{%
 \label{Fig1:second}
 \includegraphics[width=0.5\textwidth]{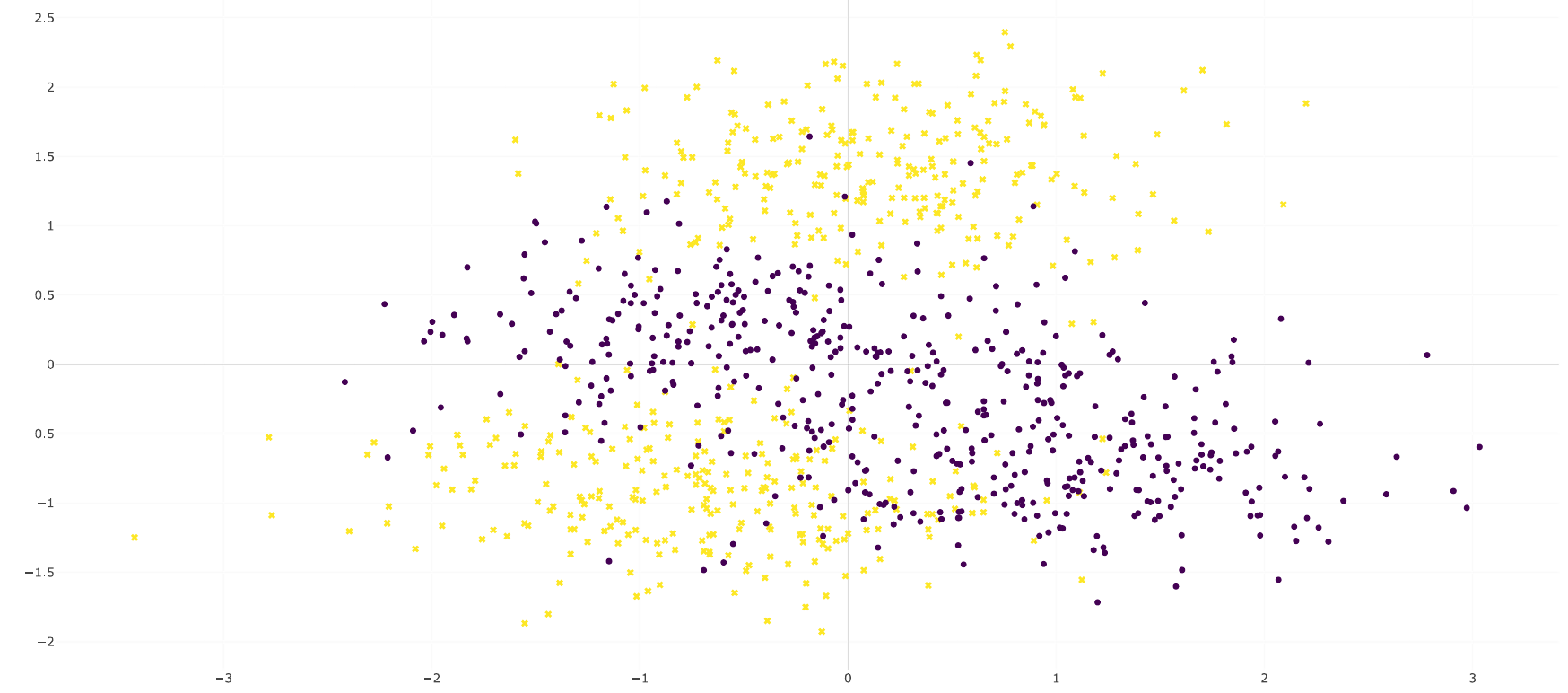}
}\\ 
\subfigure[]{%
 \label{Fig1:third}
 \includegraphics[width=0.5\textwidth]{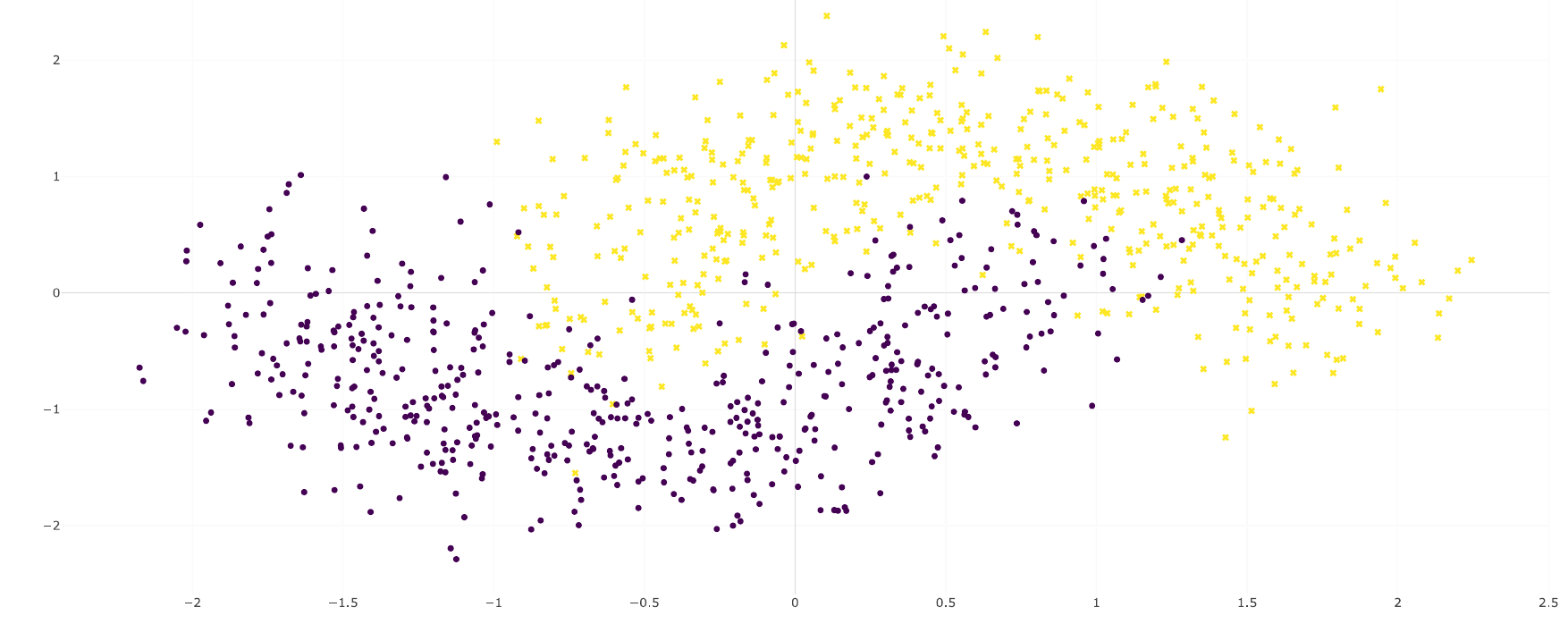}
 }%
 \subfigure[]{%
 \label{Fig1:fourth}
 \includegraphics[width=0.5\textwidth]{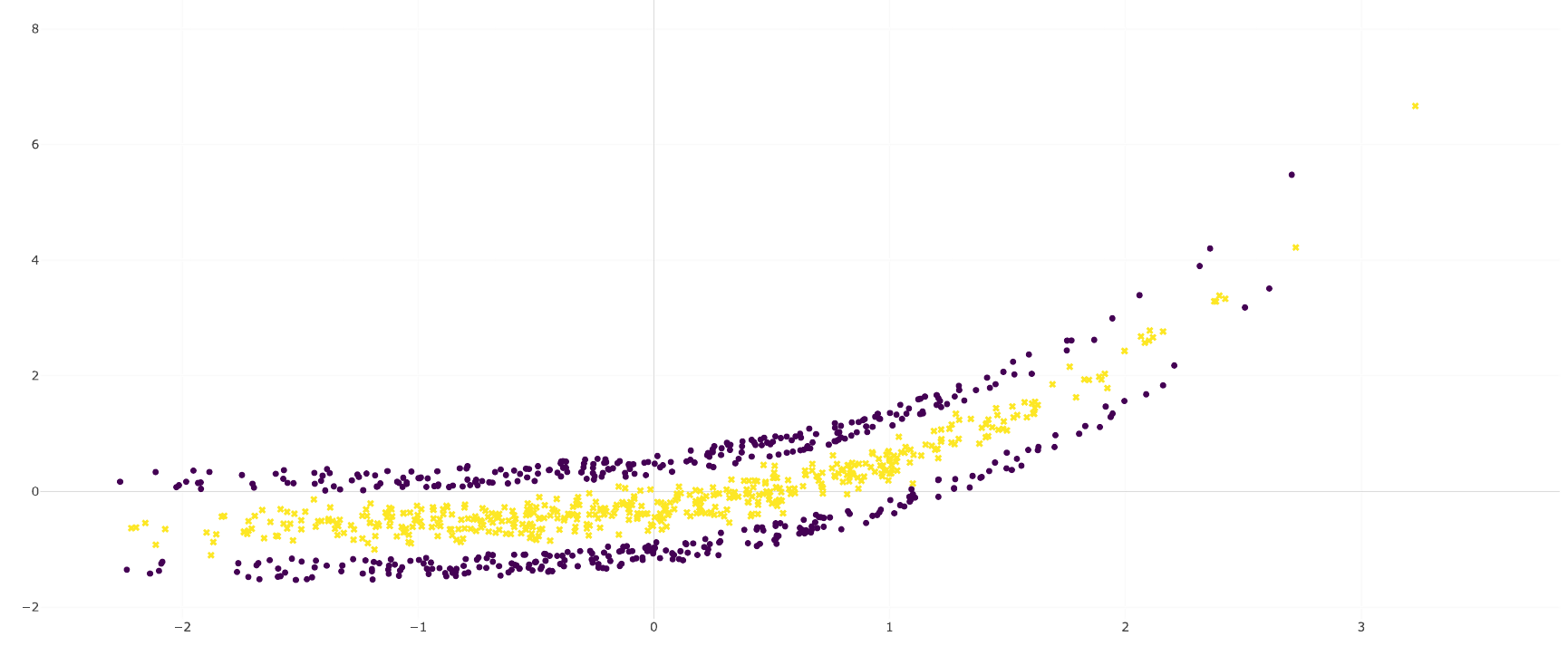}
 }\\ 
 \subfigure[]{%
 \label{Fig1:fifth}
 \includegraphics[width=0.5\textwidth]{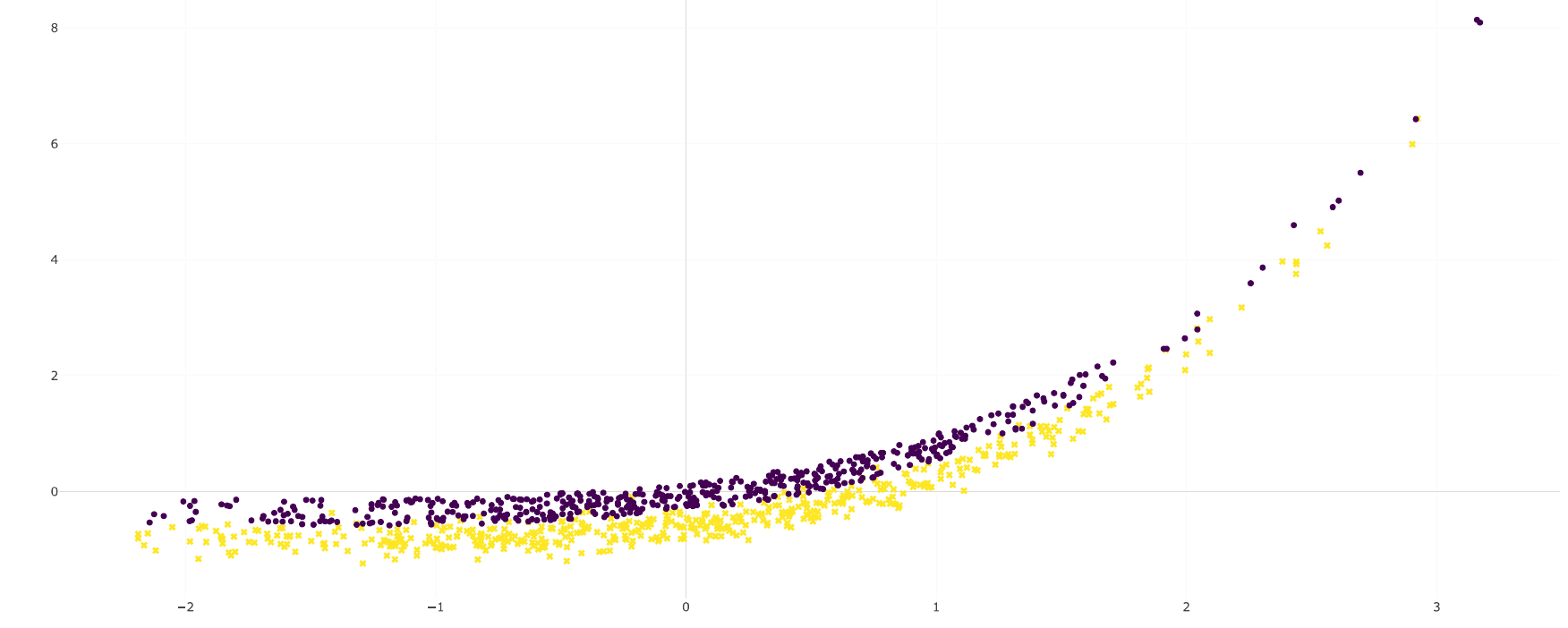}
 }%
 \end{center}
 \caption{%
Simulated patterns for the meta-learning framework.
 }%
 \label{Fig:1}
\end{figure}

Figure \ref{Fig:1} shows the five simulated patterns.

Two different sets of experiments are defined based on the five patterns for model training.
First, two-dimensional matrices of similar size (i.e., $N=1000$ and $P=2$) are considered. This meta-learning approach is referred to as CNN1.
Heterogeneous datasets in terms of the number of samples and variables are also considered: the patterns were generated using a random number of samples between 800 and 1400 and 0 to 4 irrelevant variables (Gaussian noise) were introduced, adding them to the two-dimensional patterns (i.e., $m=[800,1400]$ and $p=[2,6]$).
This approach is referred to as CNN2. The supervised learning task involves predicting the correct pattern of the datasets 
(i.e., a multiclass problem with $C=5$), and the inputs are two-class datasets, where the two classes are perfectly balanced

To highlight the performance differences of the various patterns, several well-known classifiers were used, evaluating performance on a test set of simulated tabular datasets.
The following classification methods and their respective hyperparameter values were considered:

\begin{itemize}
\item \emph{Logistic regression (Logit)}: This statistical method is suitable for a linear pattern because it constructs a separating hyperplane.
The slopes of this function are given by a vector of coefficients $\bm{\beta}$.
\item \emph{Decision tree (DT)}: This method performs a branching process in a hierarchical manner until a pruning criterion is met.
The Gini index was considered for splitting, while three different values were considered for the complexity parameter (default pruning parameter values of the `rpart' R implementation)
$cp=\{0,0.03,0.81\}$.
\item \emph{$k$-nearest neighbor ($k$-NN)}: This classifier predicts new samples according to the labels of a neighborhood of size $k$ that were obtained from the training set, with $k=\{5,7,9\}$.
\item \emph{Random forest (RF)}: This method constructs an ensemble of decision trees via bagging.
The same criteria for the DT classifier were considered, constructing 500 trees with two randomly selected variables.
\item \emph{Artificial neural network (ANN)}: A shallow ANN classifier with one hidden layer was constructed.
1, 3, and 5 hidden units were explored and the following values for the weight decay parameter: $\epsilon=\{0,0.0001,0.1\}$.
\item \emph{Support vector machine (SVM)}: The standard kernel-based SVM model with Tikhonov regularization and hinge loss was considered.
The following values for the regularization parameter were explored: $C=\{0.25,0,5,1,2,4\}$, while the kernel width $\sigma$ was held constant at a value of 1.004454.
\end{itemize}

Table \ref{tabResSim} shows the area under the curve (AUC) measure in the test set for all the methods and simulated patterns (C1 to C5).
The largest AUC value is highlighted in bold type, while the second largest AUC appears underlined.
Because all the patterns are binary classification problems, AUC is a suitable performance metric that does not rely on a threshold and can balance type-I and type-II errors adequately \cite{Hastie2009}.
\begin{table}[ht!]
\centering
\caption{Predictive performance for the five simulated patterns.}
\label{tabResSim}
\begin{tabular}{lccccc}
\toprule
 Classifier & C1 & C2 & C3 & C4 & C5 \\
\midrule
 Logit & {\bf 0.938} & 0.590 & 0.954 & 0.609 & 0.904 \\
 DT & 0.859 & 0.803 & 0.890 & 0.845 & 0.903 \\
 $k$-NN & 0.909 & {\bf 0.885} & 0.965 & \underline{0.966} & 0.970 \\
 RF & 0.910 & \underline{0.884} & \underline{0.966} & 0.962 & \underline{0.981} \\
 ANN & \underline{0.926} & 0.577 & 0.953 & 0.691 & 0.867 \\
 SVM & 0.920 & 0.882 & {\bf 0.968} & {\bf 0.981} & {\bf 0.985} \\
\bottomrule
\end{tabular}
\end{table}

Table \ref{tabResSim} shows that 
the top classification approaches vary depending on the simulated pattern:
\begin{itemize}
\item \emph{C1}: Logit is the best classifier, followed by ANN.
\item \emph{C2}: $k$-NN is the best classifier, followed by RF.
\item \emph{C3}: SVM is the best classifier, followed by RF.
\item \emph{C4}: SVM is the best classifier, followed by $k$-NN.
\item \emph{C5}: SVM is the best classifier, followed by RF.
\end{itemize}

At this stage, the proposed approach can match the expected classifiers in the sense that the logistic regression performs best for the sole linear pattern (C1), while nonlinear methods (kernel-based SVM, random forest, and $k$-NN) perform best with the remaining four nonlinear patterns.
For these patterns, the logit classifier does not perform well.
Although it is clear that ``there is no free lunch'', recommendations for classifiers can still be made if the patterns are identified correctly.
Note that the goal of training CNNs with a heterogeneous dataset in terms of size is to consider the fact that the goal is to apply the final model to real-world data, which are inherently heterogeneous.
Training with tabular datasets of different sizes (CNN2) is then a more realistic approach to meta-learning.
However, transformations are required before model training (step 2 of the proposed framework).
As mentioned in the previous section, two simple strategies were combined to manage ``images'' of different sizes (simulated tabular datasets in the proposed case): (1) PCA to shrink the datasets that are larger than a target average sample size (1099 observations for each tabular dataset, where a total of 1099 components are selected);
and (2) padding to create additional space within a dataset smaller than the target average size.
Padding is also used to achieve the target size of 7, which is the maximum size of a tabular input ${ {\textbf X},{\textbf y}}$ with the two-dimensional pattern, four irrelevant variables, and the label vector ${\bf y}$ included.
The network used is a straightforward two-stage architecture. The first stage is used for feature extraction and learning and consists of two Conv2D layers with 32 filters and a kernel size of 5x5, followed by a dropout layer with a rate of 0.25 and then two Conv2D layers with 64 filters each and a kernel size of 3x3.
The first two Conv2D layers use a kernel size of 5x5 for more general feature extraction, while the third and fourth use 3x3 kernels to extract more detailed features from the data.
The second stage performs the classification task and begins with a flattened layer, followed by a dense layer with 256 neurons, a dropout layer with a rate of 0.5, another dense layer with five neurons, and finally a softmax layer with five outputs for the five simulated patterns proposed in this experimental setting.
The network was trained with a batch size of 1000 tabular datasets and a maximum of 10 epochs, allowing early stopping.
A standard validation approach is considered, in which the validation set is used to monitor the predictive performance of the model (validation loss).
Adam was used as the optimizer, which has shown excellent empirical results in terms of efficiency, scalability, and performance (i.e., faster convergence to a minimum) \cite{Aggarwal18}.
Adam is a stochastic gradient descent method with adaptive learning rate, allowing a more effective and smooth learning process.
The following values were considered for the optimization process with Adam: $\alpha=1e-03$ (initial learning rate), $\beta_1 = 0.9$ and $\beta_2 = 0.999$ (parameters that control the exponential decay rates of the moving averages), and $\epsilon = 1e-07$ (conditioning parameter).
These parameter values were selected according to the original Adam paper \cite{kingma2017adammethodstochasticoptimization}.
The proposed methodology was implemented in the TensorFlow Python library.
Table \ref{tabExps} shows the performance of five different meta-learning strategies by considering various multiclass metrics: micro and macro averages for the precision, recall, and f1 (harmonic mean of the precision and recall).
The proposed four CNN variants and the meta-learning approach suggested in \cite{AliS1} (MF+DT) are considered, which consists of constructing several meta-features and using a decision tree for model training with the meta-features as inputs. This latter strategy includes several meta-learning studies given the set of meta-features considered in this study (see \cite{AliS1,deeva2025select}).
The heterogeneous dataset used to train CNN2 is considered to construct the meta-features.

\begin{table}[ht!]
\centering
\caption{Predictive performance for the various meta-learning approaches.}
\label{tabExps}
\begin{tabular}{lccc}
\toprule
Perf.
Measure & CNN1 & CNN2 & MF+DT \\
\midrule
precision (macro) & 0.996 & 0.986 & 0.790 \\
recall (macro) & 0.996 & 0.985 & 0.752 \\
f1-score (macro) & 0.996 & 0.985 & 0.745 \\
f1-score (micro) & 0.996 & 0.985 & 0.752 \\
\bottomrule
\end{tabular}
\end{table}

Table \ref{tabExps} shows that the two proposed CNN variants achieved similar performances with nearly perfect classification.
It can be concluded that the proposed approach can learn from various simulated patterns and identify them in other unseen tabular datasets with marginal variations.
In contrast, the two-step approach that constructs meta-features and then implements a classifier is clearly inferior in terms of its predictive capabilities, with an accuracy of approximately 75\% on average.
While Table \ref{tabExps} demonstrates the CNN models’ ability to identify known patterns in unseen simulated data, it should be interpreted with caution, as both training and validation sets share the same underlying patterns.
The true test of generalization lies in the results in Section \ref{sec:expBench}, where the model successfully recommends classifiers for real-world datasets it never encountered during training.
\subsection{Application on benchmark datasets}\label{sec:expBench}

Next, the ability of the proposed model to make suggestions for suitable classifiers is explored.
The proposed CNN2 approach is compared with the alternative strategy based on meta-features (MF+DT) on well-known binary classification datasets from the UCI Repository \cite{blake2015uci}.
CNN2 is chosen because its construction is designed to manage heterogeneous datasets, although CNN1 achieved marginally better classification results.
Also, CNN2 and MF+DT consider the same input information and are therefore comparable.
Table \ref{tabDesc} shows relevant metadata for all the benchmark datasets, including the number of examples $m$, the number of variables $n$, and the percentage of examples in each class (min.,maj.), and the imbalance ratio (IR), computed as the fraction of the majority class and minority class samples.
\begin{table}[ht!]
\centering
\caption{Descriptive statistics for all the benchmark datasets.}
\label{tabDesc}
\begin{tabular}{clcccc}
\toprule
 ID & dataset & $m$ & $n$ & \%class(min,maj) & IR \\
\midrule
 ds1 & abalone7 & 4177 & 8 & (9.3, 90.7) & 9.7 \\
 ds2 & australian-credit & 690 & 14 & (44.5,55.5) & 1.2 \\
 ds3 & banknote-auth & 1372 & 4 & (44.5,55.5) & 1.2 \\
 ds4 & breast-cancer & 569 & 30 & (37.3,62.7) & 1.7 \\
 ds5 & bupa-liver & 345 & 6 & (42, 58) & 1.4 \\
 ds6 & german-credit & 1000 & 24 & (30.0,70.0) & 2.3 \\
 ds7 & hearth-statlog & 270 & 13

& (44.4,55.6) & 1.3 \\
 ds8 & horse-colic & 300 & 27 & (33.0,67.0) & 2.0 \\
 ds9 & image-1 & 2310 & 19 & (14.3, 85.7) & 6 \\
 ds10 & image-5 & 2310 & 19 & (38.1,61.9) & 1.6 \\
 ds11 & ionosphere & 351 & 34 & (35.9,64.1) & 1.8 \\
 ds12 & monk-2 & 432 & 6 & (47.2,52.8) & 1.1 \\
 ds13 & oil-spill & 937 & 49 & (4.4,95.6) & 21.9 \\
 ds14 & phoneme & 5404 & 19 & (29.3,70.7) & 2.4 \\
 ds15 & pima-diabetes & 768 & 8 & (34.9,65.1) &

1.9 \\
 ds16 & ring & 7400 & 20 & (49.5,50.5) & 1.0 \\
 ds17 & solar-flares-M & 1389 & 10 & (4.9.95.1) & 19.4 \\
 ds18 & sonar & 208 & 60 & (46.6,53.4) & 1.4 \\
 ds19 & splice & 1000 & 60 & (48.3,51.7) & 1.1 \\
 ds20 & titanic & 2201 & 3 & (32.3,67.7) & 2.1 \\
 ds21 & waveform & 5000 & 21 & (33.1,66.9) & 2.0 \\
 ds22 & yeast02579vs368 & 1004 & 8 & (9.9, 90.1) & 9.1 \\
 ds23 & yeast5 & 1484 & 8 & (3.0, 97.0) & 32.78 \\
\bottomrule
\end{tabular}

\end{table}

The tabular datasets exhibit important differences in terms of size and imbalance ratio, which is important to assess meta-learning approaches properly (see Table \ref{tabDesc}).
Note that these datasets have no ``class'' in the sense that they have no known linear/nonlinear pattern.
Therefore, multiclass classification performance cannot be computed. However, the binary classification performance for each tabular dataset can first be computed and then compared with the patterns suggested by both CNN and MF+DT.
The results of the various classifiers on the 23 datasets are shown in Table \ref{tabResBench}.
The best classifier for each dataset is highlighted in bold type.
\begin{table}[ht!]
\centering
\caption{Predictive performance for the various benchmark datasets.}
\label{tabResBench}
\begin{tabular}{ccccccc}
\toprule
 ID & Logit & DT & $k$-NN & RF & ANN & SVM \\
\midrule
 ds1 & 0.777 & 0.82 & {\bf 0.892} & 0.864 & 0.793 & 0.822 \\
 ds2 & 0.865 & 0.863 & 0.844 & 0.865 & 0.858 & {\bf 0.866} \\
 ds3 & 0.982 & 0.977 & 0.997 & 0.986 & {\bf 1} & {\bf 1} \\
 ds4 & 0.544 & {\bf 0.644} & 0.56 & 0.613 & 0.604 & 0.5 \\
 ds5 & 0.51 & 0.594 & {\bf 0.719} & 0.671 & 0.676 & 0.702 \\
 ds6 & 0.647

& 0.514 & 0.616 & {\bf 0.666} & 0.527 & 0.617 \\
 ds7 & 0.701 & 0.755 & 0.594 & {\bf 0.79} & 0.759 & 0.775 \\
 ds8 & 0.5 & 0.519 & 0.625 & {\bf 0.745} & 0.5 & 0.5 \\
 ds9 & 0.919 & 0.96 & {\bf 0.994} & 0.987 & 0.978 & 0.981 \\
 ds10 & 0.504 & 0.556 & {\bf 0.65} & 0.611 & 0.559 & 0.536 \\
 ds11 & 0.761 & 0.786 & 0.886 & {\bf 0.935} & 0.912 & 0.932 \\
 ds12 & 0.771 & 0.822 & 0.93 & {\bf 0.941} & 0.906 &

0.899 \\
 ds13 & 0.5 & 0.5 & {\bf 0.577} & 0.5 & 0.5 & 0.5 \\
 ds14 & 0.667 & 0.753 & {\bf 0.842} & 0.804 & 0.745 & 0.792 \\
 ds15 & 0.538 & 0.498 & 0.591 & {\bf 0.634} & 0.599 & 0.585 \\
 ds16 & {\bf 0.824} & 0.82 & 0.813 & 0.821 & 0.821 & 0.821 \\
 ds17 & 0.769 & 0.817 & {\bf 0.952} & 0.929 & 0.793 & 0.826 \\
 ds18 & 0.68 & 0.681 & 0.85 & {\bf 0.771} & 0.757 & 0.68 \\
 ds19 & 0.743 & 0.723 & 0.787 &

0.783 & {\bf 0.827} & 0.807 \\
 ds20 & 0.692 & 0.675 & {\bf 0.697} & 0.677 & 0.695 & 0.695 \\
 ds21 & {\bf 0.867} & 0.803 & 0.84 & 0.862 & 0.866 & {\bf 0.867} \\
 ds22 & 0.876 & 0.93 & 0.941 & {\bf 0.946} & 0.865 & 0.913 \\
 ds23 & 0.957 & 0.976 & 0.979 & {\bf 0.981} & 0.975 & 0.972 \\
\bottomrule
\end{tabular}
\end{table}

Table \ref{tabResBench} shows that no method outperforms others in terms of performance.
In contrast to the performance of these classifiers with simulated data (see Table \ref{tabResSim}), random forest and $k$-NN achieved a better performance than SVM.
These results confirm the advantages of RF when facing noisy mixed-type data (numerical and dummy variables).
In contrast to the first stage of the framework, which follows a standard supervised scheme, a tailored strategy needs to be defined to evaluate the success of the transfer learning approach.
For this stage, the aim is to assign labels to real-world tabular datasets that do not necessarily follow a clear linear/nonlinear pattern.
The assumption is that these labels can be linked to classifiers that can perform well on these patterns.
The evaluation approach is again supervised: a classifier recommendation can be successful or not.
Success is defined as follows: For a given pattern/class C1 to C5, the two best classifiers out of a 
total of six are identified.
For a real-world dataset, the pattern is inferred with the CNN model, thus recommending these two classifiers as the best choices for it.
Next, the six classifiers are applied to the tabular dataset and the performance is evaluated, identifying the best classifier (bc).
A ``hit'' occurs when one of the two classifiers recommended by the predicted class coincides with bc.
Otherwise, the recommendation was not successful. Notice that this validation scheme differs from unsupervised approaches based on cluster purity (see, e.g., \cite{borlea2022improvement}).
Next, the results of the proposed CNN approach for meta-learning are reported in Table \ref{tabResCNN}.
For each dataset, the predicted probability of belonging to a given training pattern is shown.
The largest predicted probability for a given dataset is highlighted in bold type.
The best classifier (bc) found in Table \ref{tabResBench} is also reported, and whether the recommendation is successful or not (column ``hit'').
For ds1, for example, C2 is the predicted class with a probability of 0.978, and therefore, the recommendations for this dataset based on the simulated data are first $k$-NN and then RF (see Table \ref{tabResSim}).
The best-performing classifier is $k$-NN; therefore, the recommendation is valid.
A hit is highlighted with the symbol \ding{51}, also including the relative position of the recommended classifier (1 or 2).
In contrast, for dataset ds4, the recommendations are also $k$-NN and RF (class C2, with a confidence of 0.947), but the best classifier is DT, and therefore, the recommendation is not successful (hit=\ding{55}).
Note that there is a tie for bc in three of the 23 datasets.
In such cases, the best classifier that is considered a hit appears underlined in case one of the two best classifiers is indeed recommended by the CNN.
\begin{table}[ht!]
\centering
\caption{Predictions for the proposed CNN approach.}
\label{tabResCNN}
\begin{tabular}{cccccccc}
\toprule
 ID & C1 & C2 & C3 & C4 & C5 & bc & hit \\
 \midrule
 ds1 & 0.022 & {\bf 0.978} & 0 & 0 & 0 & k-NN & \ding{51}(1) \\
 ds2 & {\bf 0.965} & 0.026 & 0 & 0 & 0.009 & SVM & \ding{55} \\
 ds3 & 0.044 & 0 & 0 & 0.057 & {\bf 0.898} & ANN/\underline{SVM} & \ding{51}(1) \\
 ds4 & 0.041 & {\bf 0.947} & 0.001 & 0.001 & 0.01 & DT & \ding{55} \\
 ds5 & 0.163 & {\bf 0.679} & 0.02 &

0.026 & 0.111 & k-NN & \ding{51}(1) \\
 ds6 & 0.028 & 0.007 & 0.008 & 0.142 & {\bf 0.815} & RF & \ding{51}(2) \\
 ds7 & 0.09 & {\bf 0.869} & 0.009 & 0.008 & 0.025 & RF & \ding{51}(2) \\
 ds8 & 0 & 0 & 0.006 & 0.078 & {\bf 0.916} & RF & \ding{51}(2) \\
 ds9 & 0.002 & 0.15 & 0 & {\bf 0.848} & 0 & k-NN & \ding{51}(2)\\
 ds10 & 0 & 0.004 & 0 & {\bf 0.996} & 0 & k-NN & \ding{51}(2) \\
 ds11 & 0.007 & {\bf 0.992} & 0

& 0 & 0.001 & RF & \ding{51}(2) \\
 ds12 & 0.002 & {\bf 0.998} & 0 & 0 & 0 & RF & \ding{51}(2) \\
 ds13 & 0 & 0 & 0 & {\bf 0.994} & 0.006 & k-NN & \ding{51}(2) \\
 ds14 & 0 & {\bf 1} & 0 & 0 & 0 & k-NN & \ding{51}(1) \\
 ds15 & 0 & 0 & 0 & 0 & {\bf 1} & RF & \ding{51}(2) \\
 ds16 & 0 & {\bf 0.98} & 0.02 & 0 & 0 & logit & \ding{55} \\
 ds17 & 0.028 & 0.444 &

0 & {\bf 0.528} & 0 & k-NN & \ding{51}(2) \\
 ds18 & 0.126 & {\bf 0.559} & 0.075 & 0.143 & 0.097 & RF & \ding{51}(2) \\
 ds19 & 0.038 & 0.003 & 0.004 & 0.016 & {\bf 0.939} & ANN & \ding{55}\\
 ds20 & 0.008 & 0.147 & 0 & {\bf 0.845} & 0 & k-NN & \ding{51}(2) \\
 ds21 & 0 & 0.069 & {\bf 0.931} & 0 & 0 & logit/\underline{SVM} & \ding{51}(1) \\
 ds22 & 0.019 & 0.01 & {\bf 0.895} & 0.064 & 0.012 & RF & \ding{51}(2) \\
 ds23 & {\bf 0.99} &

0.008 & 0 & 0.002 & 0 & RF & \ding{55}\\
\bottomrule
\end{tabular}
\end{table}

Table \ref{tabResCNN} shows the advantages of the CNN: recommendations are successful 78.2\% of the time (18 of the 23 datasets).
Because a random recommender would achieve a 33.3\% success rate at suggesting two of the six classifiers, the proposed meta-learning approach clearly outperforms a random model.
Of the 18 hits, 13 correspond to the second-best model, while the remaining five correspond to the best model.
However, at least two classifiers should be recommended because the performance of the top method is typically similar to that of the second-best method.
Table \ref{tabResCNN} also shows the confidence of the CNN in predicting the training pattern: the largest probability is below 0.8 only in three of the 23 datasets.
This result demonstrates that the method indeed identifies one of the various patterns in the datasets.
Finally, the performance of the proposed method is compared with the traditional two-step approach.
Table \ref{tabResTree} shows the results when a DT is applied to meta-features.
The identified training pattern is highlighted with the symbol \ding{54}.
In the case of ties, the best classifier (bc) that is recommended by the meta-learning method appears underlined.
Note that landmarks are not included because they consider information that is not available for the proposed approach (e.g., the performance of the various traditional classifiers).
\begin{table}[ht!]
\centering
\caption{Predictions for the tree-based meta-learning approach trained on meta-features.}
\label{tabResTree}
\begin{tabular}{cccccccc}
\toprule
 ID & C1 & C2 & C3 & C4 & C5 & bc & hit \\
 \midrule
 ds1 & & & & & \ding{54} & k-NN & \ding{55} \\
 ds2 & & & & & \ding{54} & SVM & \ding{51}(1) \\
 ds3 & \ding{54} & & & & & \underline{ANN}/SVM & \ding{51}(2) \\
 ds4 & & & & & \ding{54} & DT & \ding{55} \\
 ds5 & & \ding{54} & & & & k-NN & \ding{51}(1) \\
 ds6 & & & & & \ding{54} & RF & \ding{51}(2) \\
 ds7 &

& & & & \ding{54} & RF & \ding{51}(2)\\
 ds8 & & & & & \ding{54} & RF & \ding{51}(2) \\
 ds9 & & & & & \ding{54} & k-NN & \ding{55}\\
 ds10 & & & & & \ding{54} & k-NN & \ding{55} \\
 ds11 & \ding{54} & & & & & RF & \ding{55} \\
 ds12 & & \ding{54} & & & & RF & \ding{51}(2) \\
 ds13 & & & & & \ding{54} & k-NN & \ding{55} \\
 ds14 & & & & \ding{54} & & k-NN & \ding{51}(2) \\
 ds15 & & & & & \ding{54} &

RF & \ding{51}(2) \\
 ds16 & \ding{54} & & & & & logit & \ding{51}(1) \\
 ds17 & & & & & \ding{54} & k-NN & \ding{55}\\
 ds18 & & & & & \ding{54} & RF & \ding{51}(2) \\
 ds19 & \ding{54} & & & & & ANN & \ding{51}(2) \\
 ds20 & & \ding{54} & & & & k-NN & \ding{51}(1) \\
 ds21 & & \ding{54} & & & & logit/SVM & \ding{55} \\
 ds22 & & & & & \ding{54} & RF & \ding{51}(2) \\
 ds23 & & & & & \ding{54} & RF & \ding{51}(2) \\
\bottomrule
\end{tabular}
\end{table}

Table

\ref{tabResTree} shows that the alternative meta-learning approach is not as successful as the proposed CNN.
The proposed method makes successful recommendations 78.2\% of the time compared to 65.2\% with the standard two-step approach (15 of the 23 datasets).
From the 15 hits, 11 correspond to the second best model.
In this sense, the alternative method is worse than the proposed CNN at identifying the best performing classifier.
\subsection{Discussion}\label{sec:expDisc}

The results in Table \ref{tabResSim} not only confirm the individual strengths of each classifier for specific types of patterns but also highlight the importance of tailoring classifier selection to the underlying data structure.
For instance, SVM consistently performs well for the more complex nonlinear patterns (C3 to C5), likely due to its ability to handle nonlinearity through kernel transformations.
In contrast, $k$-NN shows robust performance in moderately nonlinear settings, benefiting from its local decision boundaries, while RF performs strongly in mixed-noise contexts, leveraging ensemble learning to manage variability.
These findings validate our design choice to simulate diverse patterns that mirror real-world classification challenges.
Moreover, the results in Table \ref{tabExps} show the capacity of CNN-based meta-learning to internalize and generalize these pattern-specific behaviors across datasets.
The near-perfect classification accuracy for CNN1 and CNN2 indicates that the networks effectively captured the latent structure of the tabular representations and learned to associate them with the appropriate classifier profiles.
This high performance underscores the benefit of end-to-end learning, which contrasts with the hand-crafted meta-feature approach.
In particular, CNN2's robustness in handling heterogeneous dataset sizes adds practical value, as it reflects the conditions encountered in real-world applications.
Finally, the results presented in Tables \ref{tabResCNN} and \ref{tabResTree} highlight the effectiveness of the proposed CNN-based meta-learning approach for classifier recommendation.
The CNN model successfully identified one of the top two classifiers in 78.2\% of the benchmark datasets, significantly outperforming the expected success rate of a random recommender (33.3\%) and surpassing the performance of a traditional two-step meta-learning approach based on decision trees (65.2\%).
Moreover, the CNN showed high confidence in its pattern predictions, with only three datasets showing a maximum class probability below 0.8.
These findings suggest that the CNN model is not only capable of generalizing the patterns learned from synthetic data to real-world datasets but also more effective than standard meta-feature-based methods in identifying suitable classifiers.
There are some methodological challenges that can be improved. For example, more sophisticated strategies can be used to match datasets of different sizes to a CNN, such as spatial pyramid pooling (SPP).
This method consists of a special type of pooling layer that allows the inclusion of images of heterogeneous sizes.
An SPP layer pools the features in an image and generates outputs that feed the dense layers \cite{TOLOSANA2020131}.
Additionally, different CNN architectures could be explored. These challenges, however, would have a minor impact on the proposed framework because the classification accuracy is already near 100\% on the simulated datasets.
\section{Conclusions}\label{sec:Conc}

This paper proposes a novel meta-learning approach in which CNNs are used to learn directly from tabular datasets without the need to construct meta-features.
The goal is to avoid any loss of information that results from a two-step approach.
The proposed model can make accurate classifier recommendations for this type of dataset, outperforming a two-step approach based on meta-features.
The main challenge is to define a comprehensive set of simulated patterns that can be extrapolated to real-world datasets.
Results show that CNNs used in a supervised manner can achieve nearly perfect identification of simulated patterns;
however, the capacity of the proposed model to extrapolate this knowledge to real-world datasets requires further research.
Fortunately, the experiments in this study show promising results.

One limitation of the proposed approach lies in the potential risk of overfitting, particularly in the meta-learning context where the goal is to generalize knowledge across different tasks.
In the first stage of the framework—training the various meta-learning models—both CNN-based approaches achieved near-perfect classification performance.
However, in the second stage, when applying these models to real benchmark datasets, a performance drop was observed due to the presence of noise and lack of clear patterns.
Despite this, the results remain positive and indicate that the learned representations from the first stage generalize well.
To mitigate overfitting risks, especially the concern of models overfitting to simulated data, noise was introduced into the training datasets by varying sample sizes and adding irrelevant variables, which led to the more robust CNN2 model.
Another limitation is the reliance on standard CNNs for meta-feature learning, which inherently have polynomial-time complexity with respect to the input size.
Specifically, a typical CNN layer has computational complexity of $\mathcal{O}(n \cdot k^2 \cdot c \cdot m)$, where $n$ is the number of output feature maps, $k$ is the kernel size, $c$ the number of input channels, and $m$ the spatial dimension of the input.
Although the simulated tabular inputs in our case are relatively small, this complexity could become a concern in larger-scale applications.
Future work could explore efficient CNN implementations and distributed computing techniques to improve scalability and enhance the practical deployment of meta-learning strategies in more demanding settings.
This study represents an initial step toward developing a recommender system that can challenge the NFL theorem.
Therefore, this study opens interesting lines for future research. Apart from the identification of new patterns that can be simulated and subsequently linked to classifiers, the proposed method can be used to learn from more sophisticated data structures, such as text or images.
The proposed model can also be tailored to the class-imbalance problem, in which classifiers and resampling strategies can be recommended.
Finally, generative models can be considered to provide a wider diversity of training patterns.
\section*{Acknowledgements}
We gratefully acknowledge financial support from ANID, PIA/PUENTE AFB230002 and FONDECYT-Chile, grants 1250045 and 11200007. We are also grateful to the anonymous reviewers who contributed to improving the manuscript.

\bibliographystyle{abbrv}
\bibliography{biblio-meta}

\end{document}